\documentclass{article}
\usepackage{ijcai24}
\usepackage[linesnumbered,ruled,vlined]{algorithm2e}

\usepackage{setspace}
 \usepackage{multirow}
\usepackage{algorithmicx}
\usepackage{amssymb}
\usepackage{setspace}
\usepackage{amsmath}
\newcommand{\myfrac}[2]{\frac{\displaystyle #1}{\displaystyle #2}}

\usepackage{color}

\usepackage[linesnumbered, ruled]{algorithm2e}
\usepackage{algpseudocode}
\algnewcommand{\LeftComment}[1]{\Statex \(\triangleright\) #1}
\usepackage{times}
\usepackage{soul}
\usepackage{setspace}
\usepackage{url}
\usepackage[hidelinks]{hyperref}
\usepackage[utf8]{inputenc}
\usepackage[small]{caption}
\usepackage{graphicx}
\usepackage{amsmath}
\usepackage{amsthm}
\usepackage{booktabs}
\usepackage[switch]{lineno}

\title{Federated Class-Incremental Learning with New-Class Augmented Self-Distillation}

\author{
Zhiyuan Wu$^{1,2}$ \and Tianliu He$^{1,2}$ \and Sheng Sun$^{1}$ \and Yuwei Wang$^1$\thanks{\textit{(Corresponding author: Yuwei Wang)}}, \\
Min Liu$^{1,3}$
, Bo Gao$^{4}$, Xuefeng Jiang$^{1,2}$
\affiliations
$^1$Institute of Computing Technology, Chinese Academy of Sciences \\
	$^2$University of Chinese Academy of Sciences \quad $^3$Zhongguancun Laboratory \\
	$^4$Beijing Jiaotong University 
\emails
wuzhiyuan22s@ict.ac.cn \quad tianliu.he@foxmail.com \{sunsheng,ywwang,liumin\}@ict.ac.cn \\
bogao@bjtu.edu.cn \quad jiangxuefeng21b@ict.ac.cn
}

\begin{document}

\maketitle

\begin{abstract}
Federated Learning (FL) enables collaborative model training among participants while guaranteeing the privacy of raw data. Mainstream FL methodologies overlook the dynamic nature of real-world data, particularly its tendency to grow in volume and diversify in classes over time. This oversight results in FL methods suffering from catastrophic forgetting, where the trained models inadvertently discard previously learned information upon assimilating new data. In response to this challenge, we propose a novel Federated Class-Incremental Learning (FCIL) method, named \underline{Fed}erated \underline{C}lass-Incremental \underline{L}earning with New-Class \underline{A}ugmented \underline{S}elf-Di\underline{S}tillation (FedCLASS). 
The core of FedCLASS is to enrich the class scores of historical models with new class scores predicted by current models and utilize the combined knowledge for self-distillation, enabling a more sufficient and precise knowledge transfer from historical models to current models. 
Theoretical analyses demonstrate that FedCLASS stands on reliable foundations, considering scores of old classes predicted by historical models as conditional probabilities in the absence of new classes, and the scores of new classes predicted by current models as the conditional probabilities of class scores derived from historical models. Empirical experiments demonstrate the superiority of FedCLASS over four baseline algorithms in reducing average forgetting rate and boosting global accuracy. Our code is available at \url{https://github.com/skylorh/FedCLASS}.

\end{abstract}

\section{Introduction}
With the growing concern about personal data security, collaborative model training with privacy guarantees has become a substantial trend in recent years. Federated Learning (FL) \cite{yang2019federated}, an emerging distributed Machine Learning (ML) paradigm, enables multiple data owners to jointly train ML models without sharing raw data. As FL not only preserves the data privacy of participants but also aligns with the tightening landscape of data protection regulations \cite{regulation2016regulation}, it has drawn considerable interest from both academia \cite{he2020fedml} and industrial communities \cite{hard2018federated,liu2021fate}. 
Nevertheless, prevailing FL methods \cite{mcmahan2017communication,li2020federated,reddi2021adaptive} typically assume static datasets across participants, overlooking the inherently dynamic characteristics of data volume expansion and diversification that are typical in real-world scenarios.
This dynamic can lead to "catastrophic forgetting" \cite{masana2022class}, where the continuous assimilation of new information causes the model to overwrite previously acquired knowledge.
Such forgetting drastically undermines the models' proficiency in tasks that were previously mastered, thereby reducing overall performance and effectiveness \cite{zhou2023deep}.

\begin{figure}[t]
	\centering
	\includegraphics[width=0.50\textwidth]{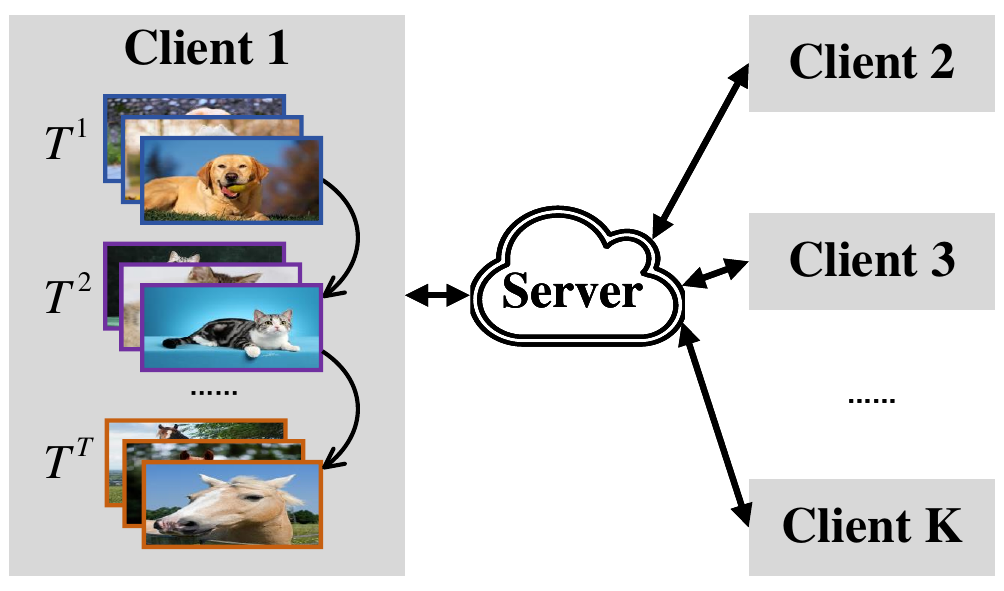}
	\caption{An overview of federated class-incremental learning. The private data of clients arrives continuously according to a series of incremental tasks, each introducing additional data that incorporates new classes.}
	\label{fcil}
\end{figure}

To counter this issue, Federated Class-Incremental Learning (FCIL) \cite{dong2022federated} is proposed to empower federated models to assimilate data with new classes while preserving their proficiency in recognizing old classes, as illustrated in Figure \ref{fcil}.
Within FCIL, self-distillation \cite{zhang2019your} is a promising technique for handling catastrophic forgetting, as it retains information about old classes by distilling knowledge from softmax-normalized outputs (called class scores) extracted by historical models into current models. However, previous methods \cite{dong2022federated,dong2023no,liu2023fedet} have shown inadequacies in reconciling the logits of old classes with those of new classes. Specifically, they simply overlay the logits from historical models onto those from current models \cite{liu2023fedet,dong2022federated,dong2023no}, which potentially results in inaccurate optimization due to scale discrepancies, hindering models from achieving optimal performance. Furthermore, the absence of reliable theoretical support hampers the further extensions of existing methods.


In this paper, we introduce a novel FCIL method, named \underline{Fed}erated \underline{C}lass-Incremental \underline{L}earning with New-Class \underline{A}ugmented \underline{S}elf-Di\underline{S}tillation (FedCLASS), which is tailored for catastrophic forgetting in FL with expanding data volumes and classes over time. 
FedCLASS leverages the scores of new classes to enrich the class scores derived from historical models, without introducing scale discrepancies among the two scores. More explicitly, the class scores of new classes extracted from current models are adopted in a manner that is mindful of their scale to reconstruct historical models' outputs that lack structural information of new classes. This process augments the breadth of knowledge that is transferred to current models via self-distillation, thereby enriching the global recognition capabilities of models on clients.
Theoretical proof indicates that FedCLASS reasonably models scores of old classes predicted by historical models as conditional probabilities where new classes are absent. Furthermore, it considers the prediction of new classes with current models as the conditional probabilities of class scores derived from historical models that assume new classes are absent. This theoretical framework solidifies the reliable foundations of FedCLASS's design.
\textbf{To our best knowledge, FedCLASS is the first federated class-incremental learning method with theoretical support.}
Our proposed method enables more holistic and precise knowledge transfer on clients, effectively mitigating catastrophic forgetting while boosting global accuracy in the face of continuously expanding data and class diversity.

In conclusion, the key contributions of our paper are as follows:
\begin{itemize}
    \item 
    We propose FedCLASS, a novel FCIL method that mitigates catastrophic forgetting by harmonizing new class scores with the outputs of historical models during self-distillation.
    \item 
    We provide theoretical analyses for FedCLASS that conform to the soundness of FedCLASS's design.
    \item 
    We conduct extensive empirical experiments on four datasets with two class-incremental settings. Results demonstrate that FedCLASS substantially reduces the average forgetting rate and markedly enhances global accuracy compared with state-of-the-art methods.
\end{itemize}

\section{Related Work}
\subsection{Federated Class-Incremental Learning}
The research field of FCIL is still in an early stage. Pioneering work \cite{dong2022federated} identifies catastrophic forgetting as the primary challenge in FCIL, mirroring a significant hurdle similarly encountered in standalone class-incremental learning \cite{masana2022class}. 
What distinguishes FCIL as a more challenging problem than its standalone counterpart is the necessity to cope with heterogeneous data distributions across clients, which significantly amplifies the difficulty of model training \cite{dong2022federated}.
By now, the most influential FCIL methods are Global-Local Forgetting Compensation (GLFC) \cite{dong2022federated} and its extension \cite{dong2023no}. These works employ customized local loss and introduce proxy servers, addressing catastrophic forgetting in FCIL from the perspectives of both clients and the server. In addition, \cite{liu2023fedet} introduces an enhancer distillation method, which aims to re-balance the acquisition of new knowledge against existing knowledge in FCIL. \textcolor{black}{However, these works have not sufficiently resolved the integration of old and new logits, which is primarily due to scale discrepancies during the distillation process in FCIL. In addition, they fall short of providing theoretical analyses.}

\subsection{Federated Learning with Knowledge Distillation}
Knowledge distillation is gaining increasing popularity for addressing a variety of challenges within FL \cite{mora2022knowledge,wu2023survey}. 
Distillation-based FL methods that conduct client-side optimization using server-side downloaded knowledge can not only accommodate model heterogeneity \cite{li2019fedmd,wu2023fedict,wu2022exploring} but demonstrate enhanced communication efficiency \cite{itahara2023distillation,wu2023fedcache}. 
Furthermore, some FL methods adopt online distillation \cite{anil2018large} as an interactive protocol among models, allowing for training larger models at locations with sufficient computational resources than those hosted on clients  \cite{deng2023hierarchical,wu2023agglomerative}. In addition, employing self-distillation as a constraint for client-side model optimization can effectively handle client drift \cite{yao2024fedgkd}, contending with noisy labels \cite{jiang2022towards}, and mitigating catastrophic forgetting \cite{lee2022preservation}. However, the aforementioned methods are not applicable in FCIL scenarios, owing to their underlying assumption that the classes of private data remain static over time.

\section{Problem Definition}
Suppose $K$ clients collaboratively train a classification model in FL, each client $k \in \{1,2,... ,K\}$ holds a private dataset ${\mathcal{D}^k}$ with corresponding local model parameters $W^k$. 
The goal of FL is to train a shared global model $W^S$ under the coordination of the central server. The overall objective is to minimize the global loss function $J^S(\cdot)$, which is the expected loss across clients, each weighted by their proportion of local sample numbers, that is:
\begin{equation}
    \mathop {\min }\limits_{{W^S}} {J^S}({W^S}) = \mathop E\limits_{k \in \{ 1,2,...K\} } \myfrac{{{N^k}}}{{\sum\nolimits_{i = 1}^K {{N^i}} }} \cdot {J^k}({W^k}),
\end{equation}
where $N^k$ is the local sample number in $\mathcal{D}^k$, $J^k(\cdot)$ is the loss function for client $k$. Unlike conventional FL setups, the volume and class diversity of private data dynamically expand in FCIL scenarios. Data is organized according to $T$ distinct tasks $\{\mathcal{T}^1,\mathcal{T}^2,...,\mathcal{T}^T\}$, where each task introduces additional data that encompasses new classes. For client $k$, the incremental data for task $\mathcal{T}^t$ is denoted as $\mathcal{D}^k_t$, and we have $\mathcal{D}^k = \bigcup_{t=1}^T \mathcal{D}_t^k$.

\textcolor{black}{Limited by computation and storage resources, when client $k$ encounters a new task $\mathcal{T}^t$, it is constrained to select only $m$ samples from the preceding $t-1$ tasks to be used for training on the current task, where $m$ (called memory size) is significantly smaller than the overall sample count accumulated by task $\mathcal{T}^t$, i.e. $m << \sum\limits_{i = 1}^t {|{\cal D}_t^k|}$. The subset of data maintained by client $k$ from the previous tasks is denoted as $\mathcal{D}^k_{t-old}$, and it is combined with the incremental data $\mathcal{D}^k_t$ for training on the current task $\mathcal{T}^t$.}



\begin{figure}[t]
	\centering
	\includegraphics[width=0.50\textwidth]{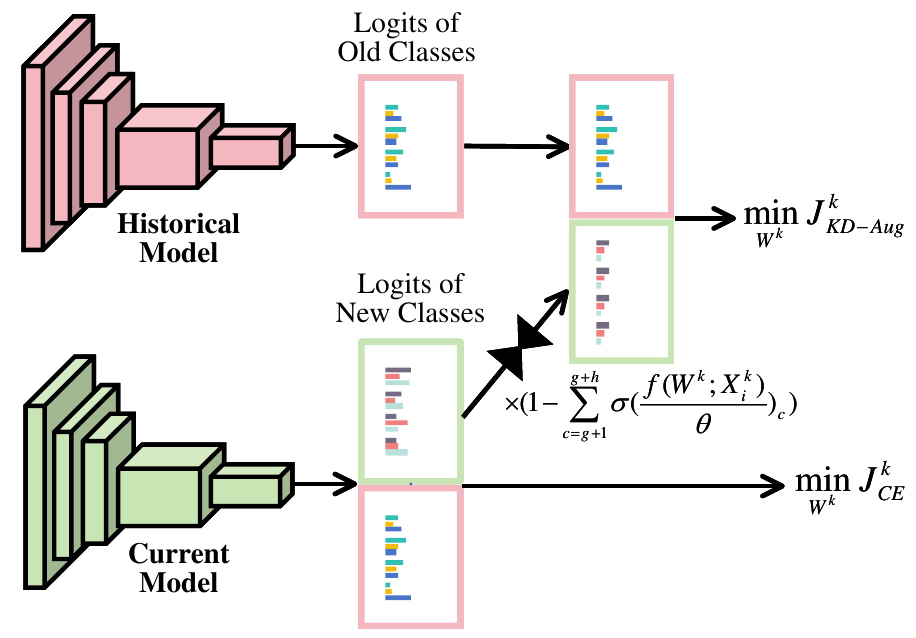}
	\caption{Illustration of new-class augmented self-distillation on client $k$.}
	\label{distill}
\end{figure}

\section{Methodology}
\subsection{New-Class Augmented Self-Distillation in Federated Class-Incremental Learning}
In the context of FL, self-distillation considers the predictions \textcolor{black}{from the global model} in previous iterations (referred to as historical models) as soft labels to guide the training of current models. 
This technique enhances the generalization capability of current models and alleviates the problem of catastrophic forgetting. Within the standard FL paradigm, the optimization objective of client $k$, predicated on compliance with the traditional server-side model aggregation protocol \cite{mcmahan2017communication}, can be formulated as follows:
\begin{equation}
\mathop {\min }\limits_{{W^t}} {J^k}({W^k}) = \mathop {\min }\limits_{{W^t}} [J_{CE}^k({W^k}) + \beta  \cdot J_{KD}^k({W^k})],
\end{equation}
subject to:
\begin{equation}
    {J_{CE}^k({W^k}) = \mathop {{\rm{ }}E}\limits_{(X_i^k,y_i^k) \in {\mathcal{D}^k}} {L_{CE}}(\sigma (f({W^k};X_i^k)),y_i^k)},
\end{equation}
\begin{equation}
    \begin{array}{l}
\;\;\;\;J_{KD}^k({W^k})\\
 = \mathop {{\rm{ }}E}\limits_{(X_i^k,y_i^k) \in {\mathcal{D}^k}} {L_{KL}}(\sigma (f({W^k};X_i^k))||\sigma (\myfrac{{f(W_{old}^k;X_i^k)}}{\theta }))
\end{array},
\end{equation}
where $\sigma(\cdot)$ is the softmax function, $L_{CE}(\cdot)$ is the cross-entropy loss, $L_{KL}(\cdot)$ is the Kullback-Leibler divergence loss, and $\beta$ the coefficient weighting the distillation loss term. $W^k_{old}$ refers to the parameters of historical model on client $k$, $f(W^*;\cdot)$ describes a nonlinear mapping instantiated by neural network parameters $W^*$, and $\theta$ represents the temperature hyper-parameter for distillation.
However, in the FCIL scenario, the number of old classes is smaller than the total number of classes accumulated to the current task.
This difference can result in an information gap during self-distillation, specifically a deficit of probability information for new classes within the pattern of historical models.

To tackle this issue, FedCLASS utilizes the class scores derived from current models to approximate historical models' predictions for new classes, as illustrated in Figure \ref{distill}. This ensures a comprehensive provision of class information for self-distillation over current models. Without loss of generality, we assume that within the class scores of the current task are of $g+h$ dimensions. The initial $g$ dimensions pertain to old classes, and the subsequent $h$ dimensions pertain to new classes. 
We set the scale of historical models' new class scores to align with those of current models. Meanwhile, the scores for the old classes from historical models are scaled to ensure a normalized total, where the sum of scores among all classes is equal to 1.
Accordingly, the historical model's predicted class score $z_i^k$ for the $i$-th sample in $\mathcal{D}_{t - \text{old}}^k \cup \mathcal{D}_t^k$ is transferred to the current model on client $k$, as expressed by the following formula:
\begin{equation}
\label{zij}
    \begin{array}{*{20}{l}}
\begin{array}{l}
{(z_i^k)_j} = \\
\left\{ {\begin{array}{*{20}{l}}
{\sigma {{(\myfrac{{f({W^k};X_i^k)}}{\theta })}_j},g < j \le g + h}\\
{\sigma {{(\myfrac{{f(W_{old}^k;X_i^k)}}{\theta })}_j} \cdot (1 - \sum\limits_{c = g + 1}^{g + h} {\sigma {{(\myfrac{{f({W^k};X_i^k)}}{\theta })}_c}} )}\\{ \; \; \; \; \; \; \; \; \; \; \; \; \; \; \; \; \; \; \; \; \; \; \; \; \; \; \; \; \; \; \; \;  \; \; \; \; \; \; \; \; \; \; \; \; \; \; \; \; \; \; \; \; \; \; \; \; \; \;0 < j \le g}
\end{array}} \right.
\end{array}\\
\end{array},
\end{equation}
subject to:
\begin{equation}
    {(X_i^k,y_i^k) \in \mathcal{D}_{t - old}^k \cup \mathcal{D}_t^k}.
\end{equation}
Accordingly, client $k$ establishes a novel optimization objective $J^k_{Aug}(\cdot)$ for its current model based on new-class augmented self-distillation, that is: 
\begin{equation}
    \mathop {\min }\limits_{{W^t}} J_{Aug}^k({W^k}) = \mathop {\min }\limits_{{W^t}} [J_{CE}^k({W^k}) + \beta  \cdot J_{KD - Aug}^k({W^k})],
\end{equation}
subject to:
\begin{equation}
  \begin{array}{l}
 \; \; \; \; J_{CE}^k({W^k}) \\= \mathop E\limits_{(X_i^k,y_i^k) \in \mathcal{D}_{t - old}^k \cup \mathcal{D}_t^k} {L_{CE}}(\sigma (f({W^k};X_i^k)),y_i^k),
  \end{array}
\end{equation}
\begin{equation}
\begin{array}{l}
\; \; \; \; J_{KD - Aug}^k({W^k})\\
 = \mathop {{\rm{ }}E}\limits_{(X_i^k,y_i^k) \in \mathcal{D}_{t - old}^k \cup \mathcal{D}_t^k} {L_{KL}}(\sigma (f({W^k};X_i^k))||z_i^k).
\end{array}
\end{equation}

We formulate our proposed FedCLASS as Algorithms \ref{alg1} and \ref{alg2}. 
\textcolor{black}{In the following section, we prove that FedCLASS effectively models scores of old classes predicted by historical models as conditional probabilities where new classes are absent. Moreover, the prediction of new classes with current models is considered as the conditional probabilities of class scores derived from historical models that assume new classes are absent.}
Established on the aforementioned sound assumptions, our analyses substantiate the proposed FedCLASS with a reliable theoretical guarantee.



\begin{algorithm}[t]
\newcommand{\removelatexerror}{\let\@latex@error\@gobble}
	\caption{FedCLASS on Client $k$}
	\SetAlgoNoLine
	\LinesNumbered

	\label{alg1}
	\begin{spacing}{1.3}
        \begin{algorithmic}[1]
            \State \textbf{for} each task $\mathcal{T}^t \in [\mathcal{T}^1, \mathcal{T}^2,...,\mathcal{T}^T]$:
            \State \quad \textbf{for} each client $k \in \{1,2,...,K\}$ in parallel:
            \State \quad \quad \textbf{if} $\mathcal{T}^t = \mathcal{T}^1$:
            \State \quad \quad \quad optimize $J_{CE}^k(W^k)$
            \State \quad \quad \textbf{else}:
            \State \quad \quad \quad $\mathcal{D}^k_{t-old} \leftarrow Subset(\mathcal{D}_{(t-1) - old}^k \cup \mathcal{D}_t^k)$
            \State \quad \quad \quad $W_{old}^k \leftarrow W^k$
            \State \quad \quad \quad extend $W^k$ to fit new classes
            \State \quad \quad \quad \textbf{while} $\mathcal{T}^t$ not finished:
            \State \quad \quad \quad \quad optimize $J_{Aug}^k(W^k)$
            \State \quad \quad \quad \quad upload $W^k$ to the server
            \State \quad \quad \quad \quad update $W^k$ with downloaded $W^S$
           \end{algorithmic}
	\end{spacing}
\end{algorithm}

\subsection{Theoretical Analyses of FedCLASS}
\noindent
\textbf{Definition 1.} In the context of the current task $\mathcal{T}^t$, we define the set of $g$ old classes as $\{C^{old}_1,C^{old}_2, ... ,C^{old}_g\} = \textbf{C}^{old}$, and define the set of $h$ new classes as $\{C^{new}_{g+1},C^{new}_{g+2},... ,C^{new}_{g+h}\} = \textbf{C}^{new}$.

\noindent
\textbf{Definition 2.} Let $S^{old}$ represent the historical state, and $S^{new}$ represent the current state. Therefore, $p(\cdot|S^{old})$ and $p(\cdot|S^{new})$ signify the conditional probabilities in the historical and current states, respectively.

\noindent
\textbf{Assumption 1.} \textit{(Class Complementarity).} 
Any class is either new or old, i.e.
\begin{equation}
\label{comp}
    {{\bf{C}}^{old}} = \overline {{{\bf{C}}^{new}}} .
\end{equation}

\noindent
\textbf{Assumption 2.}  \textit{(Conditional Probability Modeling).} 
Given any sample ${(X_i^k,y_i^k) \in \mathcal{D}_{t - old}^k \cup \mathcal{D}_t^k} $ as input, \textcolor{black}{the class score of any old class $C^{old}_u \in \textbf{C}^{old}$ predicted by historical models on client $k$ is modeled as the conditional probability of the old class with the absence of the new classes, which is represented by the $u$-th dimension of the softmax output derived from historical models,} that is:
\begin{equation}
{p(C_u^{old}|{S^{old}}) = p(C_u^{old}|{{\bf{C}}^{new}},{S^{old}}),}
\end{equation}
\begin{equation}
    {p(C_u^{old}|{{\bf{C}}^{new}},{S^{old}}) = \sigma {{(\myfrac{{f(W_{old}^k;X_i^k)}}{\theta })}_u},}
\end{equation}
subject to:
\begin{equation}
{u \in \{ 1,2,...,g\} }.
\end{equation}

\begin{algorithm}[t]
\newcommand{\removelatexerror}{\let\@latex@error\@gobble}
	\caption{FedCLASS on the Server}
	\SetAlgoNoLine
	\LinesNumbered
	\label{alg2}
	\begin{spacing}{1.3}
        \begin{algorithmic}[1]
            \State \textbf{for} each task $\mathcal{T}^t \in [\mathcal{T}^1, \mathcal{T}^2,...,\mathcal{T}^T]$:
            \State \quad \textbf{while} task $\mathcal{T}^t$ not finished:
            \State \quad \quad waiting for $K$ clients upload $W^k$, $k \in \{1,2,...,K\}$
            \State \quad \quad ${W^S} \leftarrow \sum\limits_{k = 1}^K {\myfrac{{|\mathcal{D}_t^k| \cdot {W^k}}}{{\sum\limits_{l = 1}^K {|\mathcal{D}_t^l|} }}} $
            \State \quad \quad broadcast ${W^S}$ to $K$ clients
           \end{algorithmic}
	\end{spacing}
\end{algorithm}

\noindent
\textbf{Assumption 3.} \textit{(Probability Approximation).}
Given any sample ${(X_i^k,y_i^k) \in \mathcal{D}_{t - old}^k \cup \mathcal{D}_t^k} $ as input, the class score of any new class $C^{new}_u \in \textbf{C}^{new}$ predicted by historical models on client $k$ is approximated by the class score of current models on new classes, which is represented by the $u$-th dimension of the softmax output derived from current models, that is:
\begin{equation}
\label{as3-1}
{p(C_u^{new}|{S^{old}}) = p(C_u^{new}|{S^{new}}),}
\end{equation}
\begin{equation}
\label{as3-2}
{p(C_u^{new}|{S^{new}}) = \sigma {{(\myfrac{{f({W^k};X_i^k)}}{\theta })}_u},}
\end{equation}
subject to:
\begin{equation}
\label{as3-3}
{u \in \{ g + 1,g + 2,...,g + h\}. }
\end{equation}

\noindent
\textbf{Theorem.} FedCLASS is equivalent to modeling the following two items during the self-distillation process: 1) modeling scores from old classes derived from historical models as conditional probabilities in the absence of new classes. 2) modeling the prediction of new classes with current models as the conditional probabilities of class scores derived from historical models that assume new classes are absent.


\noindent
\textbf{Proof.} 
To prove the proposition is equivalent to solving for the class score $z^k_i$ to be transferred to current models under Definitions 1-2 as well as Assumptions 1-3, and to confirm that the inferred $z^k_i$ is equivalent to that presented in Eq. (\ref{zij}). 
Specifically, $z^k_i$ is composed of two components within the historical state: the first is the conditional probability of any new class, denoted as $p(C_{u}^{new}|{S^{old}}),u\in\{g+1,g+2,... ,g+h\}$; the second is the conditional probability of any old class, denoted as $p(C_{u}^{old}|{S^{old}}),u\in\{1,2,... ,g\}$.

At first, we solve for $p(C_{u}^{new}|{S^{old}}),u\in\{g+1,g+2,... ,g+h\}$. According to Eq. (\ref{as3-1}), Eq. (\ref{as3-2}) and Eq. (\ref{as3-3}), we have:
\begin{equation}
\begin{array}{l}
p(C_u^{new}|{S^{old}}) = p(C_u^{new}|{S^{new}}) = \sigma {(\myfrac{{f({W^k};X_i^k)}}{\theta })_u},\\
\end{array}
\end{equation}
subject to:
\begin{equation}
    u \in \{ g + 1,g + 2,...,g + h\}.
\end{equation}

Next, we solve for $p(C_{u}^{old}|{S^{old}})$ as follows:
\begin{equation}
\begin{array}{*{20}{l}}
\;\;\;\; \; {p(C_u^{old}|{S^{old}})}\\
\begin{array}{l}
 = p(C_u^{old}|{{\bf{C}}^{new}},{S^{old}}) \cdot p({{\bf{C}}^{new}}|{S^{old}})\\
 \;\;\;\; + p(C_u^{old}|\overline {{{\bf{C}}^{new}}} ,{S^{old}}) \cdot p(\overline {{{\bf{C}}^{new}}} |{S^{old}}).
\end{array}
\end{array}
\end{equation}
As
\begin{equation}
    p(C_u^{old}|{{\bf C}^{new}},{S^{old}}) = \myfrac{{p(\{C_u^{old}\} \cap {{\bf C}^{new}}|{S^{old}})}}{{p({{\bf C}^{new}}|{S^{old}})}},
\end{equation}
according to Eq. (\ref{comp}), we can infer that:
\begin{equation}
    {{\bf C}^{{old}}} \cap {{\bf C}^{new}}=\emptyset.
\end{equation}
Thus, for any ${C_u^{{old}}}\in {{\bf C}^{{old}}}$, we have:
\begin{equation}
    {C_u^{{old}}}\notin {{\bf C}^{{new}}}.
\end{equation}
Hence,
\begin{equation}
    {\{ C_u^{{old}}\}}\cap {{\bf C}^{new}}=\emptyset.
\end{equation}
Thereinto,
\begin{equation}
\begin{array}{*{20}{l}}
\; \; \; \;{p(C_u^{old}|{{\bf{C}}^{new}},{S^{old}})}\\
{ = \myfrac{{p(\{ C_u^{old}\}  \cap {{\bf{C}}^{new}}|{S^{old}})}}{{p({{\bf{C}}^{new}}|{S^{old}})}}}\\
{ = \myfrac{{p(\emptyset |{S^{old}})}}{{p({C^{new}}|{S^{old}})}}}\\
{ = 0}
\end{array}.
\end{equation}
Accordingly,
\begin{equation}
    {p(C_u^{old}|{S^{old}}) = p(C_u^{old}|{\overline{{\bf C}^{new}}},{S^{old}}) \cdot p(\overline{{\bf C}^{new}}|{S^{old}})}.
\end{equation}
On the one hand,
\begin{equation}
    p(C_u^{old}|\overline {{{\bf{C}}^{new}}} ,{S^{old}})=  \sigma {(\myfrac{{f({W^k_{old}};X_i^k)}}{\theta})_u}.
\end{equation}
On the other hand,
\begin{equation}
\begin{array}{*{20}{l}}
\;\;\;\;{p(\overline {{{\bf{C}}^{new}}} |{S^{old}})}\\
{
 = 1 - p({{\bf{C}}^{new}}|{S^{old}})}\\
{ = 1 - p({{\bf{C}}^{new}}|{S^{new}})}
\\
{ = 1 - \sum\limits_{c = g + 1}^{g + h} {p(C_c^{new}|{S^{new}})} }\\
{ = 1 - \sum\limits_{c = g + 1}^{s + h} {\sigma {{(\myfrac{{f({W^k};X_i^k)}}{\theta })}_c}} }
\end{array}.
\end{equation}
Therefore, 
\begin{equation}
\begin{array}{l}
\;\;\;\;p(C_u^{old}|{S^{old}})\\
 = \sigma {(\myfrac{{f(W_{old}^k;X_i^k)}}{\theta })_u} \cdot (1 - \sum\limits_{c = g + 1}^{s + h} {\sigma {{(\myfrac{{f({W^k};X_i^k)}}{\theta })}_c}} ).
\end{array}
\end{equation} 
Thus, the class score $z^k_i$ to be transferred from historical models to current models can be formulated as:
\begin{equation}
    {(z_i^k)_j} = \left\{ \begin{array}{l}p(C_u^{old}|{S^{old}}),0 < u \le g\\p(C_u^{new}|{S^{old}}),g < u \le g + h\end{array} \right.,
\end{equation}
that is:
\begin{equation}
\begin{array}{l}
{(z_i^k)_j} = \\
\left\{ {\begin{array}{*{20}{l}}
{\sigma {{(\myfrac{{f({W^k};X_i^k)}}{\theta })}_u},g < u \le g + h}\\
{\sigma {{(\myfrac{{f(W_{old}^k;X_i^k)}}{\theta })}_u} \cdot (1 - \sum\limits_{c = g + 1}^{s + h} {\sigma {{(\myfrac{{f({W^k};X_i^k)}}{\theta })}_c}} ),}\\ \; \; \; \; \; \; \; \; \; \; \; \; \; \; \; \; \; \; \; \; \; \; \; \; \; \; \; \; \; \; \; \;  \; \; \; \; \; \; \; \; \; \; \; \; \; \; \; \; \; \; \; \; \; \; \; \; \; \;0 < u \le g
\end{array}} \right.
\end{array},
\end{equation}
which is equivalent to Eq. (\ref{zij}). The theorem is proved. $\hfill \square$

\begin{table*}[!t]
\caption{Performance comparisons between FedCLASS and baseline algorithms on three datasets with two incremental tasks. Global accuracy is the model accuracy over all classes.
\textbf{Bold} numbers represent the best performance measured by global accuracy (\%) or averaging forgetting rate (\%). The same as below.}
        \setlength{\tabcolsep}{6pt}
	\renewcommand\arraystretch{1.25}
	\centering
 \label{two-task}
\begin{tabular}{c|l|cc|cc}
\hline
\textbf{Dataset}                   & \multicolumn{1}{c|}{\textbf{Method}} & \textbf{$\mathcal{T}^1$ Accuracy} & \textbf{$\mathcal{T}^2$ Accuracy} & \textbf{Global Accuracy} & \textbf{(Average) Forgetting Rate} \\ \hline
\multirow{5}{*}{\textbf{SVHN}}     & FedAvg                               & 26.56                & 95.26                & 62.38                    & 66.38                              \\
                                   & GLFC                                 & 26.31                & 92.96                & 62.28                    & 49.93                              \\
                                   & FedICARL                             & 31.89                & 94.75                & 64.22                    & 61.19                              \\
                                   & FedWA                                & 41.08                & 93.79                & 69.34                    & 51.86                              \\
                                   & \textbf{FedCLASS}                     & 76.72                & 92.59                & \textbf{86.14}           & \textbf{16.22}                     \\ \hline
\multirow{5}{*}{\textbf{CIFAR-10}} & FedAvg                               & 34.66                & 90.44                & 62.55                    & 41.02                              \\
                                   & GLFC                                 & 18.68                & 82.54                & 50.61                    & 48.60                              \\
                                   & FedICARL                             & 33.93                & 89.72                & 61.82                    & 40.78                              \\
                                   & FedWA                                & 33.68                & 88.44                & 61.06                    & 42.00                              \\
                                   & \textbf{FedCLASS}                     & 46.00                & 87.04                & \textbf{66.52}           & \textbf{29.68}                     \\ \hline
\multirow{5}{*}{\textbf{SYN-NUM}}  & FedAvg                               & 34.66                & 99.39                & 89.72                    & 19.27                              \\
                                   & GLFC                                 & 41.78                & 98.86                & 70.69                    & 56.59                              \\
                                   & FedICARL                             & 74.02                & 99.16                & 86.79                    & 25.08                              \\
                                   & FedWA                                & 70.51                & 98.94                & 84.89                    & 28.51                              \\
                                   & \textbf{FedCLASS}                     & 95.03                & 98.91                & \textbf{97.00}           & \textbf{3.99}                      \\ \hline
\end{tabular}

\vspace{15pt}

\caption{Performance comparisons between FedCLASS and four baseline algorithms on two datasets with three incremental tasks.}
        \setlength{\tabcolsep}{6.5pt}
	\renewcommand\arraystretch{1.25}
	\centering
 \label{three-task}
\begin{tabular}{c|l|ccccc|cc}
\hline
\textbf{Dataset}                   & \multicolumn{1}{c|}{\textbf{Method}} & \textbf{$\mathcal{T}^1$ Acc.} & \textbf{$\mathcal{T}^2$ Acc.} & \textbf{$\mathcal{T}^3$ Acc.} & \textbf{\begin{tabular}[c]{@{}c@{}}$\mathcal{T}^2$ Forg.\\ Rate\end{tabular}} & \textbf{\begin{tabular}[c]{@{}c@{}}$\mathcal{T}^3$ Forg.\\ Rate\end{tabular}} & \textbf{\begin{tabular}[c]{@{}c@{}}Global\\ Acc.\end{tabular}} & \textbf{\begin{tabular}[c]{@{}c@{}}Avg. Forg.\\ Rate\end{tabular}} \\ \hline
\multirow{5}{*}{\textbf{EMNIST}}   & FedAvg                               & 58.34            & 68.72            & 95.95            & 37.17                                                            & 30.98                                                            & 73.65                                                          & 34.08                                                              \\
                                   & GLFC                                 & 0.00             & 20.32            & 71.48            & 24.80                                                            & 59.32                                                            & 29.30                                                          & 42.06                                                              \\
                                   & FedICARL                             & 59.90            & 68.47            & 95.93            & 44.70                                                            & 29.42                                                            & 74.13                                                          & 37.06                                                              \\
                                   & FedWA                                & 27.35            & 67.78            & 94.82            & 39.29                                                            & 45.57                                                            & 61.79                                                          & 42.43                                                              \\
                                   & \textbf{FedCLASS}                     & 56.29            & 73.57            & 95.87            & 16.24                                                            & 27.86                                                            & \textbf{74.44}                                                 & \textbf{22.05}                                                     \\ \hline
\multirow{5}{*}{\textbf{CIFAR-10}} & FedAvg                               & 40.78            & 30.83            & 95.97            & 39.98                                                            & 50.96                                                            & 54.35                                                          & 45.47                                                              \\
                                   & GLFC                                 & 18.70            & 7.97             & 95.43            & 52.37                                                            & 71.33                                                            & 38.50                                                          & 61.85                                                              \\
                                   & FedICARL                             & 41.60            & 32.17            & 96.33            & 43.68                                                            & 51.02                                                            & 55.19                                                          & 47.35                                                              \\
                                   & FedWA                                & 40.45            & 29.00            & 93.63            & 40.33                                                            & 36.92                                                            & 52.97                                                          & 38.63                                                              \\
                                   & \textbf{FedCLASS}                     & 49.75            & 24.07            & 94.07            & 15.42                                                            & 41.45                                                            & \textbf{55.34}                                                 & \textbf{28.44}                                                     \\ \hline
\end{tabular}
\end{table*}

\section{Experiments}
\subsection{Implementation Details}
We conduct simulation experiments on a physical server equipped with multiple NVIDIA GeForce RTX 3090 GPUs and enough memory capacity. Our experiments encompass four datasets: SVHN \cite{netzer2011reading}, CIFAR-10 \cite{krizhevsky2009learning}, SYN-NUM \cite{ganin2015unsupervised}, and EMNIST \cite{cohen2017emnist}. 
We adopt the data partitioning strategy from FedML \cite{he2020fedml}, setting the hyperparameter $\alpha=0.5$ to simulate non-independent and identically distributed data distributions across 20 clients.
To evaluate the effectiveness of FedCLASS in multiple class-incremental learning scenarios, we devise the following experimental settings: 
\begin{itemize}
    \item 
    \textbf{Two-task Class-Incremental Learning.} We establish a class-incremental learning scenario on SVHN and CIFAR-10 datasets \textcolor{black}{comprising two sequential tasks}, introducing five classes in each phase, represented as $\{5,5\}$.
    \item 
    \textbf{Three-task Class-Incremental Learning.} We establish a class-incremental learning setting on the EMNIST dataset with $\{17,15,15\}$. On the CIFAR-10 dataset, we set $\{4,3,3\}$.
\end{itemize}
Specifically, for three-task class-incremental learning on CIFAR-10, we set the distillation weighting coefficient $\beta$ to 8. For the remaining experiments, $\beta$ is set as 5. The memory size $m$ is set as \{20, 50, 50\} for experiments on SVHN, CIFAR-10, and SYN-NUM datasets, respectively. For the EMNIST dataset, $m$ is set to be twice the total number of classes (2 memory for each class) in every task.

Regarding baseline algorithms, we compare FedCLASS not only with the standard FL algorithm, FedAvg \cite{mcmahan2017communication}, but also with class-incremental learning methods designed for both federated and standalone contexts, which are GLFC \cite{dong2022federated}, ICARL \cite{rebuffi2017icarl}, and WA \cite{zhao2020maintaining}. For GLFC, \textcolor{black}{we focus on its client-side loss function}. 
For ICARL and WA, we integrate their standalone training processes with the aggregation strategy introduced by FedAvg, thus formulating the composite algorithms FedICARL and FedWA, respectively.
In terms of model architectures and optimization techniques, all clients adopt the ResNet8 \cite{he2016deep} model structure. Besides, we utilize the SGD optimizer for all experiments, setting the learning rate to $0.01$, momentum to $0.9$, weight decay value to $1\times 10^{-5}$, and batch size to $32$.

Regarding experimental criteria, we systematically report the performance of each algorithm on individual tasks, evaluating them using two key measures: task-specific accuracy and forgetting rate \cite{liu2020mnemonics}. Furthermore, we present overarching metrics, global accuracy and average forgetting rate, as two primary indicators for evaluating the overall performance of algorithms.

\begin{table*}[]
\caption{Performance of FedCLASS under varying $\beta$ configurations. $\Delta$ represents the change in the average forgetting rate relative to the setting \textcolor{black}{presented in the first line.} The same as below.}
        \setlength{\tabcolsep}{8pt}
	\renewcommand\arraystretch{1.25}
	\centering
        \label{impact-beta}
\begin{tabular}{c|ccccc|cc|c}
\hline
$\beta$ & \textbf{$\mathcal{T}^1$ Acc.} & \textbf{$\mathcal{T}^2$ Acc.} & \textbf{$\mathcal{T}^3$ Acc.} & \textbf{\begin{tabular}[c]{@{}c@{}}$\mathcal{T}^2$ Forg.\\ Rate\end{tabular}} & \textbf{\begin{tabular}[c]{@{}c@{}}$\mathcal{T}^3$ Forg.\\ Rate\end{tabular}} & \textbf{\begin{tabular}[c]{@{}c@{}}Global\\ Acc.\end{tabular}} & \textbf{\begin{tabular}[c]{@{}c@{}}Avg. Forg.\\ Rate\end{tabular}} & \textbf{$\Delta$} \\ \hline
$0$             & 58.34            & 68.72            & 95.95            & 37.17                                                            & 30.98                                                            & 73.65                                                          & 34.08                                                              & -              \\
$1$             & 54.10            & 70.08            & 96.23            & 29.29                                                            & 31.39                                                            & 72.65                                                          & 30.34                                                              & 3.74$\downarrow$          \\
$3$             & 52.29            & 71.48            & 95.98            & 20.97                                                            & 31.38                                                            & 72.36                                                          & 26.18                                                              & 7.90$\downarrow$          \\
$5$             & 56.29            & 73.57            & 95.87            & 16.24                                                            & 27.86                                                            & 74.44                                                          & 22.05                                                              & 12.03$\downarrow$         \\
$8$             & 60.85            & 76.82            & 95.45            & 11.60                                                            & 22.81                                                            & 76.99                                                          & 17.21                                                              & 16.87$\downarrow$         \\ \hline
\end{tabular}

\vspace{15pt}

\caption{Performance of FedCLASS under varying $m$ configurations.}
        \label{impact-m}
        \setlength{\tabcolsep}{6pt}
	\renewcommand\arraystretch{1.25}
	\centering
     \label{three-tasks}
 \begin{tabular}{c|ccccc|cc|c}
\hline
$m$ & \textbf{$\mathcal{T}^1$ Acc.} & \textbf{$\mathcal{T}^2$ Acc.} & \textbf{$\mathcal{T}^3$ Acc.} & \textbf{\begin{tabular}[c]{@{}c@{}}$\mathcal{T}^2$ Forg.\\ Rate\end{tabular}} & \textbf{\begin{tabular}[c]{@{}c@{}}$\mathcal{T}^3$ Forg.\\ Rate\end{tabular}} & \textbf{\begin{tabular}[c]{@{}c@{}}Global\\ Acc.\end{tabular}} & \textbf{\begin{tabular}[c]{@{}c@{}}Avg. Forg.\\ Rate\end{tabular}} & \textbf{$\Delta$} \\ \hline
$0$          & 9.18             & 32.97            & 96.22            & 30.56                                                            & 72.11                                                            & 44.55                                                          & 51.34                                                              & -              \\
$2|{{\bf C}^{new}} \cup {{\bf C}^{old}}|$          & 56.29            & 73.57            & 95.87            & 16.24                                                            & 27.86                                                            & 74.44                                                          & 22.05                                                              & 29.29$\downarrow$         \\
$4|{{\bf C}^{new}} \cup {{\bf C}^{old}}|$          & 66.29            & 80.63            & 95.43            & 11.90                                                            & 19.15                                                            & 80.17                                                          & 15.53                                                              & 35.81$\downarrow$         \\
$6|{{\bf C}^{new}} \cup {{\bf C}^{old}}|$          & 70.15            & 83.35            & 94.83            & 10.38                                                            & 15.65                                                            & 82.24                                                          & 13.02                                                              & 38.32$\downarrow$         \\
$8|{{\bf C}^{new}} \cup {{\bf C}^{old}}|$          & 72.94            & 84.62            & 94.22            & 7.79                                                             & 13.37                                                            & 83.46                                                          & 10.58                                                              & 40.76$\downarrow$         \\ \hline
\end{tabular}
\end{table*}

\subsection{Performance Comparison}
Tables \ref{two-task} and \ref{three-task} present the performance comparisons between FedCLASS and four baseline algorithms with two and three incremental tasks, respectively. From a holistic perspective, FedCLASS consistently outperforms FedAvg, GLFC, FedICARL, and FedWA across all experimental configurations. This is evidenced by its notably superior global accuracy and reduced average forgetting rates. These results indicate the advanced capability of FedCLASS to alleviate catastrophic forgetting and enhance the overall model performance. 
\textcolor{black}{It is worth noting that FedCLASS does not consistently outperform all baseline algorithms in every individual task.} 
When considering task-specific accuracy, there exists a benchmark algorithm that surpasses FedCLASS in $\mathcal{T}^2$ over two-task class-incremental learning, as well as in $\mathcal{T}^1$ and $\mathcal{T}^3$ over three-task class-incremental learning. In terms of task-specific forgetting rate, FedCLASS exhibits a more significant forgetting in $\mathcal{T}^3$ compared to FedWA over three-task class-incremental learning. The isolated underperformance is due to the trade-off between preserving old class information and learning new classes. Some baselines only focus on learning new classes, displaying better results on new tasks at the cost of severe forgetting of historical information, adversely affecting their global accuracy. \textcolor{black}{However, these isolated instances do not detract from the overall superiority of FedCLASS, as performances on different tasks are commonly trade-offs in FCIL.} 
The state-of-art results in average performance are sufficient to demonstrate the robust balance of FedCLASS between preserving established knowledge and assimilating new information, which is of critical importance in FCIL contexts.

\subsection{Ablation Studies}
We conduct ablation studies to investigate the impacts of the distillation coefficient $\beta$ and the memory size $m$ on global accuracy as well as the average forgetting rate. The ablation experiments maintain the same setting as that adopted on three-task class-incremental learning experiments over the EMNIST dataset. Exclusively, the hyper-parameters $\beta$ and $m$ are fine-tuned in this section to examine their effects.
Table \ref{impact-beta} presents the performance of FedCLASS under varying $\beta$ configurations. As displayed, the average forgetting rate consistently decreases with the increase of the value of $\beta$, declining from 34.08\% to 17.21\%. 
This substantial reduction verifies the beneficial effect of self-distillation in reducing catastrophic forgetting. 
Moreover, the global accuracy exhibits a non-linear characteristic in response to the varying $\beta$ values, initially decreasing before rebounding as $\beta$ continued to rise. 
This pattern suggests exploring the critical equilibrium between acquiring new knowledge and preserving old information, which is essential for optimizing overall performance.
Table \ref{impact-m} presents the performance of FedCLASS under varying $m$ configurations. 
It can be observed that both global accuracy and average forgetting rates can be improved with larger memory sizes. These improvements are pronounced, with global accuracy climbing from 44.55\% to 83.46\%, and the average forgetting rate declining from 51.34\% to 10.58\%. 
These findings underscore that the expansion of memory sizes under resource availability can further enhance the precision obtained by FedCLASS and reduce the degree of catastrophic forgetting.

\section{Conclusion and Future Work}
In this paper, we propose FedCLASS, a novel federated class-incremental learning method based on new-class augmented self-distillation. FedCLASS aims to mitigate catastrophic forgetting in the context of federated learning with progressively increasing data volume and classes. 
Our proposed method augments the \textcolor{black}{class scores of historical models with new class scores from current models,} and utilizes the hybrid knowledge for self-distillation, which facilitates a more comprehensive and precise knowledge transfer from historical models to current models. 
Theoretical analyses confirm \textcolor{black}{the reliable foundations} upon which FedCLASS is designed. Experiments validate the state-of-the-art performance of FedCLASS in reducing average forgetting rate and improving global accuracy.
\textcolor{black}{Future work will focus on implementing FedCLASS over larger numbers of tasks, and a particular emphasis will be placed on tackling long-term forgetting.}

\section*{Acknowledgments}
We thank Jinda Lu from University of Science and Technology of China and Yuhan Tang from Beijing Jiaotong University for inspiring suggestions.

\bibliographystyle{named}

\end{document}